\pdfoutput=1
\documentclass[11pt]{article}

\usepackage[final]{acl}

\usepackage{times}
\usepackage{latexsym}

\usepackage[T1]{fontenc}
\usepackage[utf8]{inputenc}

\usepackage{microtype}

\usepackage{inconsolata}

\usepackage{graphicx}
\usepackage{enumitem}
\usepackage{multirow}
\usepackage{hhline}
\usepackage{adjustbox}
\usepackage{booktabs}
\usepackage{tabularx}
\usepackage{amsmath}
\usepackage{float}

\newcolumntype{C}{>{\centering\arraybackslash}X}

\title{OASIS: Order-Augmented Strategy for Improved Code Search}

\author{
  \textbf{Zuchen Gao $^{1,2}$, Zizheng Zhan $^{3}$, Xianming Li $^{1}$, Erxin Yu $^{1}$} \\ 
  \textbf{Ziqi Zhan, Haotian Zhang $^{3}$, Bin Chen $^{3}$, Yuqun Zhang $^{2}$, Jing Li $^{1}$\thanks{Jing Li is the corresponding author.}}  \vspace{0.5em} \\
  $^{1}$ \includegraphics[width=0.36cm]{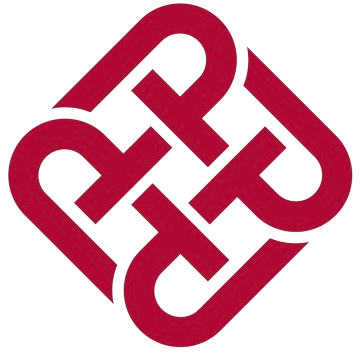} The Hong Kong Polytechnic University \\
  $^{2}$ \includegraphics[width=0.36cm]{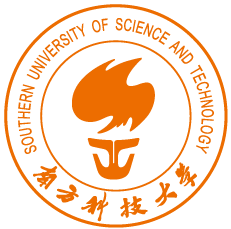} Southern University of Science and Technology, $^{3}$ \includegraphics[width=0.36cm]{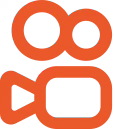} Kuaishou Technology. \\
  \texttt{\{zuchen.gao, xianming.li, erxin.yu\}@connect.polyu.hk} \\ \texttt{jing-amelia.li@polyu.edu.hk}  \ \ \ \texttt{zhangyq@sustech.edu.cn} \\
  \texttt{zhanziqi934@gmail.com,} \texttt{\{zhanzizheng, zhanghaotian, chenbin\}@kuaishou.com} \\
  \includegraphics[width=0.36cm]{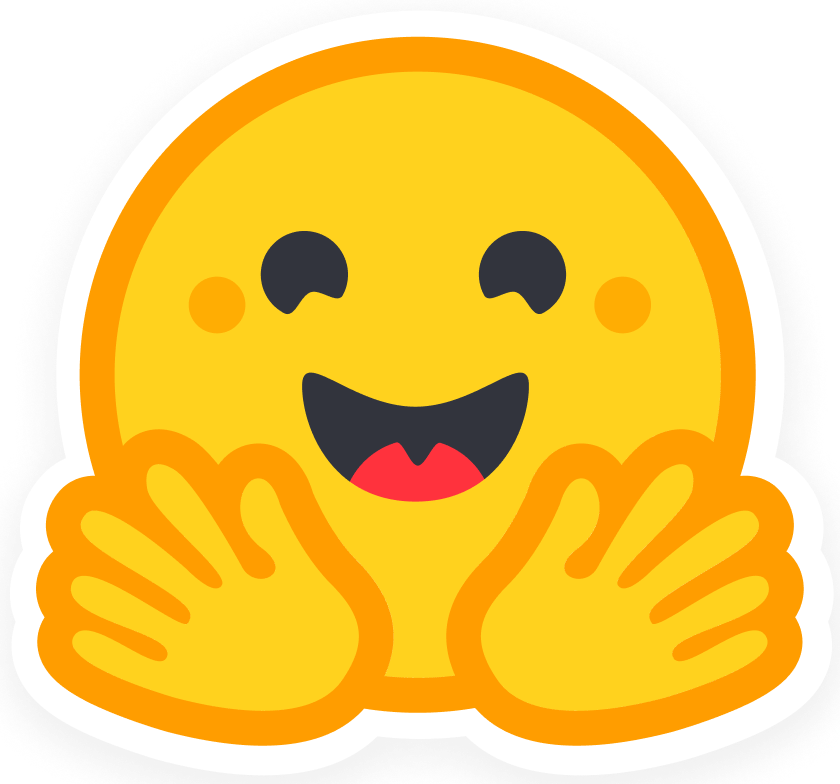} \texttt{\small \url{https://huggingface.co/Kwaipilot/OASIS-code-embedding-1.5B}}
}

\begin{document}
\maketitle
\begin{abstract}
Code embeddings capture the semantic representations of code and are crucial for various code-related large language model (LLM) applications, such as code search.
%Traditional training approaches for the code embedding model 
Previous training primarily relies on optimizing the InfoNCE loss by comparing positive natural language (NL)-code pairs with in-batch negatives.
%optimizing high-quality positive pairs of natural language code descriptions and code snippets using InfoNCE loss with in-batch negative optimization. 
However, due to the sparse nature of code contexts, training solely by comparing the \emph{major} differences between positive and negative pairs may fail to capture deeper semantic nuances.
%the major difference between the positive pair and the in-batch negative sample may result in superficial code embedding
%, and the potential of mined negative pairs for training in code embedding remains unexplored. 
To address this issue, we propose a novel \underline{o}rder-\underline{a}ugmented \underline{s}trategy for \underline{i}mproved code \underline{s}earch (OASIS). 
%fine-grained similarity labels for code pairs, providing 
It leverages order-based similarity labels to train models to capture \emph{subtle} differences in similarity among negative pairs.
%based on datasets such as CoSQA, AdvTest, and CodeSearchNet 
Extensive benchmark evaluations demonstrate that our OASIS model significantly outperforms previous state-of-the-art models focusing solely on major positive-negative differences.
It underscores the value of exploiting subtle differences among negative pairs with order labels for effective code embedding training.
%This evidence indicates that OASIS can generate high-quality code embeddings and highlights the effectiveness of leveraging fine-grained similarity signals from negative pairs for improved code search performance.

%adopting the OASIS method can produce high-quality code embeddings and benefit downstream code search tasks.

% \input{latex/abstract_xm}

\end{abstract}

\section{Introduction}
Code search tasks aim to retrieve the code snippet that best matches a given natural language (NL) query, thereby significantly enhancing developer productivity. 
A common approach is to leverage \textit{code embedding} vectors to represent the semantics of the code for measuring its similarity to NL queries in code-NL matching tasks \cite{nie2016query, husain2019codesearchnet, shuai2020improving, parvez2021retrieval, zeng2022extensive, di2023code}. Building on LLM advancements, code embeddings have significantly benefited recent code applications, including Retrieval-Augmented Generation (RAG) \cite{asai2023retrieval, gao2023retrieval} and code completion (ReAcc) \cite{lu2022reacc, tan2024prompt} and repairing \cite{xiang2024far}.

Code embedding models are typically trained with contrastive learning. 
Here, the model learns embeddings of NL queries or codes by pulling semantically similar positive pairs closer and pushing semantically unrelated negative pairs apart \cite{shi2023cocosoda, li2022coderetriever, zhang2024code}. Each batch contains multiple positive pairs, and a sample in a positive pair forms a negative pair with another sample outside the pair. 
The training objective is to minimize the distance between positive pairs relative to all in-batch negative pairs using the \textit{InfoNCE loss function} \cite{oord2018representation}.

% within a given batch with multiple positive pairs, and 

\begin{figure}[t]
  \includegraphics[width=\columnwidth]{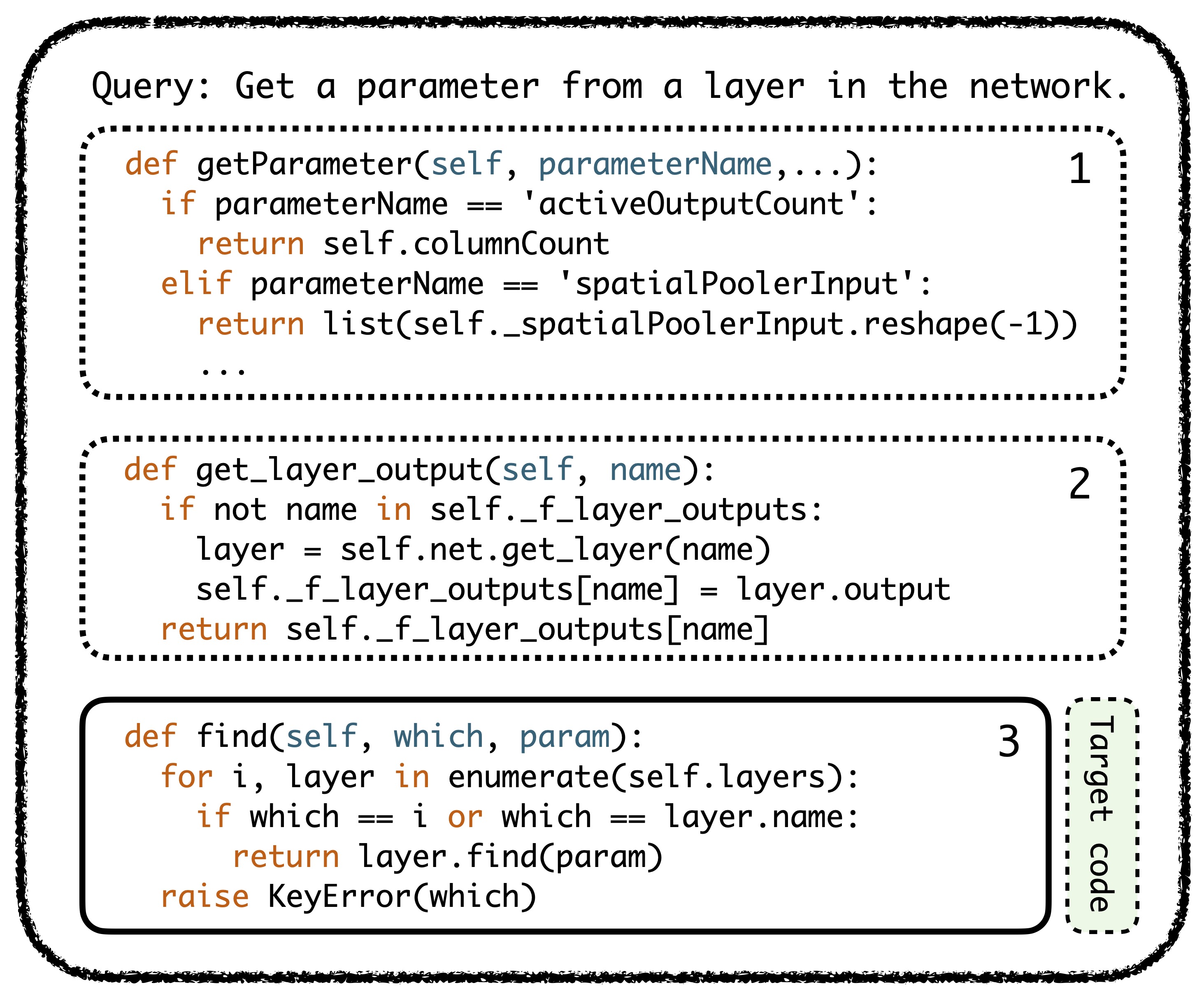}
  \vspace{-1.5em}
  \caption{
  An example of NL query and its three candidate code snippets from the CSN Python dataset. Snippets 1 and 2 exhibit higher word overlap (i.e., superficial similarity), yet Snippet 3 is the correct target.
  %A case from model w/o order-based data on CSN Python. Results of top3 candidate code snippets are presented for query: "Get a parameter from a layer in the network." Superficially similar but functionally incorrect code 1 is ranked first, while the actual target code, which lacks lexical overlap, is ranked at third.
  }
  \vspace{-1 em}\label{fig:intro_case}
\end{figure}

However, most previous work relies on the difference between positive and negative pairs (i.e., \emph{major} difference). 
This approach may result in superficial semantics due to the \emph{sparse} nature of code context, where even a subtle change can lead to significant variations in functionality and meaning.
To illustrate this insight, we present a sample from the CSN Python dataset in Figure \ref{fig:intro_case}.
Focusing solely on major differences may lead to matching the query with code snippets 1 and 2, which only exhibit superficial similarity through word overlap. Detailed elaboration can be found in appendix \ref{sec:motivation_case_elaboration}.

To address this concern, we aim to research how to exploit \emph{subtle} differences among negative pairs to improve code search. 
Prior work has primarily focused on intra-differences among negative pairs in two ways: some models have investigated ``hard'' negative pairs, which are difficult for models to differentiate \cite{karpukhin2020dense, gao2021scaling, zhang2021adversarial}, while others have adopted weighted optimization objectives to automatically balance learning from both hard and easy negatives \cite{li2022soft, li2023rethinking, zhuang2024not}. Details of existing methods for hard negative samples can be found in appendix \ref{sec:hard_negative_analysis}.
In contrast, we propose \textbf{OASIS} (\underline{O}rder-\underline{A}ugmented \underline{S}trategy for \underline{I}mproved code \underline{S}earch), which leverages \emph{order}-based similarity labels to capture deeper semantic nuances. In this way, OASIS can better distinguish subtle differences among negative pairs through finer-grained comparisons for effective code search.

Concretely, OASIS leverages LLMs to generate high-quality docstrings for function-level code snippets, treating these docstrings as equivalent to queries \cite{li2022coderetriever, zhang2024code}. Within the same repository, it pairs these code snippets with docstrings from other functions to create highly similar negative pairs, assigning them similarity labels. 
They act as order labels, providing additional order-augmented training signals to help the model distinguish between negative pairs.
To further improve the quality of the similarity labels, a program analysis approach is employed to identify inaccurately labeled sample pairs, and an LLM is utilized to generate refined similarity labels.

Building on OASIS, we automatically synthesized a large-scale training dataset with 53 million NL-Code pairs across 9 programming languages for code search. Subsequently, high-quality code embeddings were trained on this dataset and evaluated on three widely-used code search benchmarks: AdvTest \cite{lu2021codexglue}, CodeSearchNet \cite{husain2019codesearchnet}, and CoSQA \cite{huang2021cosqa}. 
%The OASIS method can effectively identify highly similar negative pairs and can very accurately understand the subtle differences in sample pairs for natural language-code similarity learning.

The main results first show that OASIS achieved state-of-the-art (SOTA) performance across all datasets, with an average improvement of 3\% in the NL2Code tasks and 9\% in the Code2Code tasks.
These demonstrate the usefulness of capturing subtle differences among negative samples and the effectiveness of order labels in identifying them.
An ablation study then indicates that all components of OASIS contribute positively to its effectiveness. 
Finally, we further analyze OASIS outputs, uncovering its superiority in handling hard (challenging) cases and interpreting how it enhances code search.
%Finally, further experimentation revealed that OASIS possesses a robust capacity to discern highly similar negative pairs, detailed analysis and visualization are presented in the final section.

In summary, our contributions are as follows:
% \begin{itemize}[label=\textbullet, leftmargin=*, itemindent=0pt, labelwidth=10pt, labelsep=5pt]
  
  $\bullet$ To the best of our knowledge, OASIS is the first code embedding model to explore subtle differences among negative pairs using order labels.
  %OASIS model is the first model to distinguish subtle difference among negative pairs using order relationships.
  
  $\bullet$ Building on OASIS, we contributed synthesized training data with million-scale, multi-language NL-Code pairs to advance code search.
  
  %We introduced an innovative data synthesis method, OASIS, and developed a large-scale, high-quality dataset comprising 53 million NL-Code pairs across nine languages, intended to facilitate future training endeavors.
  
  $\bullet$ Extensive experimental results demonstrate that subtle differences among negative pairs are crucial for effective code embedding training.
% \end{itemize}

\section{Related Work}

OASIS is in line with previous work on code embedding, which builds upon the concept of \emph{text embedding}. 
Text embedding aims to generate high-dimensional vectors that encode the semantic representations of text based on its context. 
Many prior studies have utilized contrastive learning to investigate semantic similarity for embedding learning \cite{zhang2020unsupervised, gao2021simcse, chuang2022diffcse, zhuo2023whitenedcse}. 
Given the presence of similarity labels in the STS (Semantic Textual Similarity) datasets, many studies concentrate on utilizing this label as an additional optimization target to enhance text embedding capabilities, which proved to be effective \cite{liu2023rankcse, seonwoo2022ranking, huang2024cosent, li2023angle}.
However, code embeddings remain relatively underdeveloped due to the challenges of similarity labeling compared to text, attributed to the \emph{sparse} context of code. Our work aims to mitigate the gap.
%However, the field of code lacks similarly detailed similarity-labeled datasets, rendering these refined training methodologies challenging to implement.

OASIS is also related to broader code-related tasks. 
Inspired by the NLP paradigm, the typical practice involves representation learning through pre-training on a large code corpus, followed by fine-tuning for specific downstream tasks \cite{feng2020codebert, tan2024llm4decompile}. Here, \emph{code embedding} is a crucial pre-training task, aiming to align code semantics with NL queries for code search. Following text embedding, they employ contrastive learning to explore code similarity \cite{guo2022unixcoder, shi2023cocosoda, zhang2024code}. Some also incorporated code structures, such as data flow or Abstract Syntax Trees (AST), as auxiliary objectives \cite{guo2020graphcodebert, guo2022unixcoder}. 

However, most previous work relies on \emph{major} positive-negative differences, ignoring \emph{subtle} differences among negative pairs crucial for learning code semantics in sparse contexts.
While some explored weighted training to emphasize harder-to-distinguish negative pairs \cite{li2022coderetriever, li2022soft, li2023rethinking},  OASIS proposed order labels to enable finer-grained comparisons to explore subtle differences.

%Contrastive learning is also adopted as a standard procedure for code embedding model training \cite{guo2022unixcoder, shi2023cocosoda, zhang2024code}. 
% Additionally, some works in the code domain have attempted weighted training on various negative sample pairs based on their similarity, aiming to mitigate the performance degradation caused by false negatives \cite{li2022coderetriever, li2022soft, li2023rethinking}. However, these methods generally utilize self-generated similarities during the training phase as input, which are not necessarily accurate. Moreover, the observed performance enhancements may stem from assigning lower weights to high-similarity negative pairs rather than actually improving the representation of these samples.

\section{OASIS Framework}
\label{sec:method}
\begin{figure*}[h]
    \centering
    \includegraphics[width=1\textwidth]{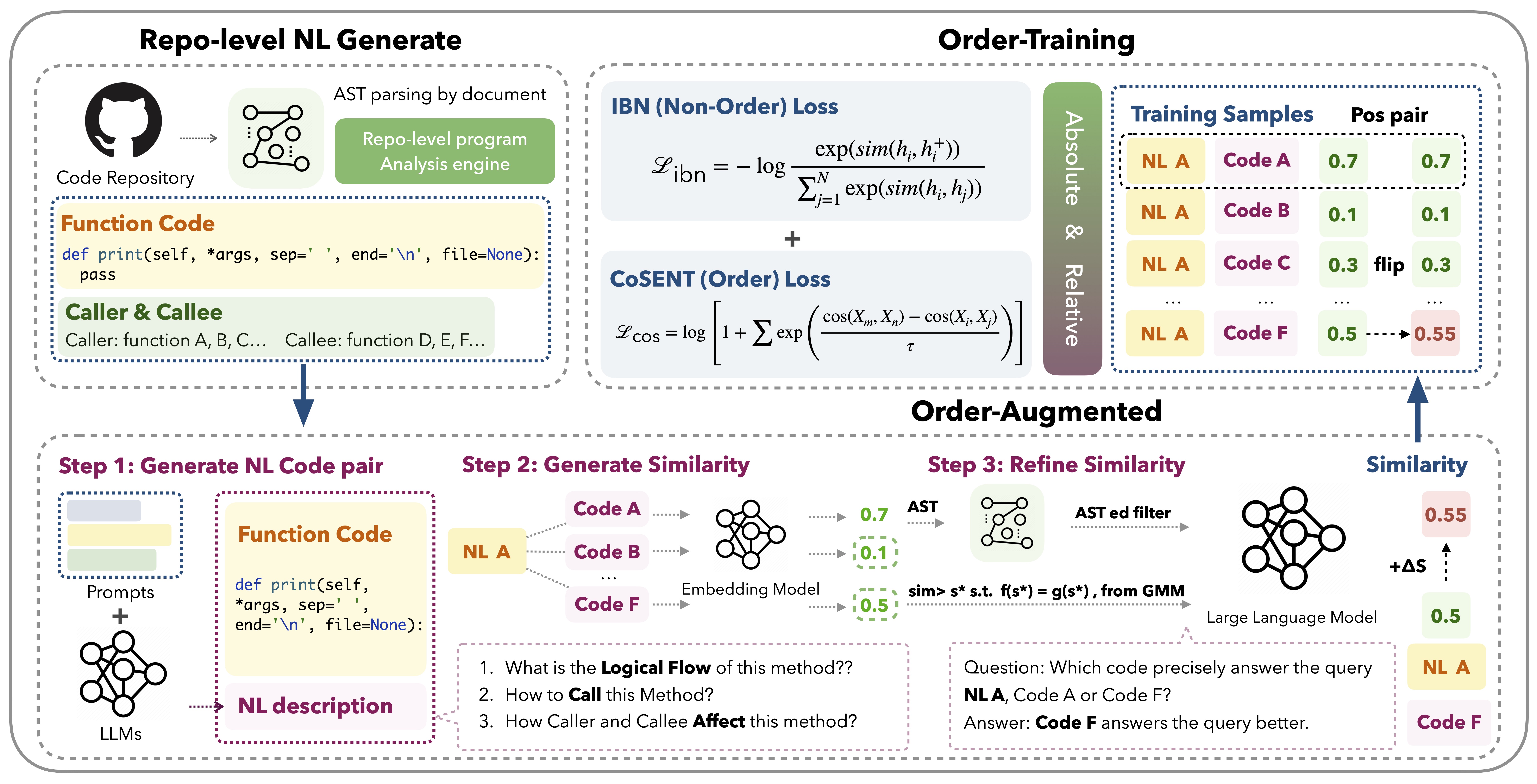}
    \caption{The overall framework of OASIS. OASIS begins by using program analysis to enhance prompts for pairing code with generated docstrings. Then, these pairs are augmented and annotated for similarity, after which suboptimal labeled negative pairs will be selected with AST and threshold strategies and similarity will be adjusted subsequently. Finally, these refined similarity labels are used in the optimization of hybrid objective.}
    \label{fig:framework}
\end{figure*}

In this section, we will introduce the fundamental framework of the OASIS method, as depicted in Figure \ref{fig:framework}. Initially, the dosctring generation and similarity generation module will be discussed in Section \ref{sec:method1}. This will be followed by an exposition on the refinement approach for similarity in Section \ref{sec:method2}. Section \ref{sec:method3} will encompass the optimization objectives for the overall training phase of OASIS.

\subsection{Similarity Annotation}
\label{sec:method1}
The essential thought of OASIS is to fully exploit the implicit information among negative sample pairs, which necessitates high-quality queries of code and high-quality sample pair similarity data. Docstrings can serve as the natural language description for code which is functionally considered semantically analogous to a query. The code and the corresponding docstring are considered as a positive pair and should be embedded to the same vector space \cite{li2022coderetriever, zhang2024code}.

The data for the OASIS method is sourced from open-source code on GitHub. To ensure the quality and consistency of the docstrings, the initial step 1 in the training methodology involves the generation of docstrings for code snippets. This process begins with program analysis at the repository level, where information about the function's callers and callees is extracted. This additional information, along with the code itself, is incorporated into a prompt to facilitate the generation of docstrings by an LLM.%, serving as the basis for subsequent similarity generation and refinement of the sample pairs.

Subsequently, the docstrings generated for all functions within a repository are utilized for data augmentation and similarity generation in step 2. Specifically, for a given docstring A, \(K\) other code snippets are randomly selected from the same repository to form negative sample pairs. The similarity labels for these negative sample pairs are then calculated using embeddings generated by another embedding model. The rationale for this approach is that code within the same repository often shares similar semantics and functions, and may even overlap lexically to a significant extent. This naturally results in a large pool of high-quality negative sample pairs, which, while similar to the original sample pairs, still exhibit subtle differences. The similarity score can help to distinguish the false negative pairs by simply assigning a high similarity. In Equation \ref{negative_pair}, \( Q \) represents the query, which is the docstring for code fragment \( i \) during training. \( N \) signifies the number of functions within a repository, and \( C^{j_k \in N \setminus \{i\}} \) denotes the procedure of randomly selecting \( K \) code snippets other than \( i \), to be paired with \( Q \) to form new sample pairs. The term \( sim \) is the similarity score calculated from embeddings, which lies in the interval \([0,1)\).

{\small
% \begin{equation}
% (C^i, Q^i,sim=1.0)
% \end{equation}
\begin{equation}
\label{negative_pair}
(C^{j_k \in N \setminus\{i\}}, Q^i,sim\in[0,1)), k \in \{1,...,K\}
\end{equation}
}

\subsection{Similarity Refinement}
\label{sec:method2}
The third step involves finely calibrating the similarity derived from the second step. The similarity labels from the second step can help the model shape an approximate similarity relationship, but utilizing data annotated with similarity scores derived from an embedding model may constrain the performance of the model below that annotation model. Therefore, precisely adjusting the similarity can further enhance performance. A `candidate pair' refers to an inaccurately annotated pair within the negative sample pairs, which are derived from a positive pair. The similarity refinement is done by selecting candidate pairs and adjusting similarity.

Two methods are employed to extract candidate pairs requiring refinement. The first method filters candidate pairs based on the similarity scores, employing Gaussian Mixture Model (GMM) to fit the distribution of similarity scores across all sample pairs. Here, the similarity scores exhibit a bimodal distribution, with one peak corresponding to positive pairs and the other to negative pairs. The intersection of these two distributions is taken as the threshold value \(s^*\) for delineating positive and negative sample pairs. The equation is as below:
\begin{equation}
    \small
    \begin{split}
    &f(x) =\frac{1}{\sqrt{2\pi \sigma_1^2}} \exp\left(-\frac{(x - \mu_1)^2}{2\sigma_1^2}\right) \\
     &= \frac{1}{\sqrt{2\pi \sigma_2^2}} \exp\left(-\frac{(x - \mu_2)^2}{2\sigma_2^2}\right) =  g(x)\\
    \end{split}
\end{equation}
where $f, g$ are the resulting distributions from GMM. If the similarity of a negative pair exceeds this threshold \(s^*\) or surpasses the similarity of the corresponding positive pair's, then the similarity score of this negative pair is likely to be inaccurate.

The other method selects candidate pairs whose code and the original code's parsed Abstract Syntax Tree (AST) have a low ratio of edit distance to the sum of the nodes of both trees, which means only a few of the nodes are required to be modified to be the other AST. This selection method compensates for candidate pairs that are structurally similar in code but dissimilar at the lexical level.

Finally, an LLM is used to determine, with the docstring as a query, whether the candidate code or the code of the positive pair better answers the query. If LLM determines the candidate code can also satisfy docstring, then similarity of candidate pair is adjusted positively with $\Delta s$, which is the optimal value attained by grid search. This approach is inspired by the robust performance of LLMs in binary choice tasks. Directly scoring pairs with an LLM typically results in most negative pairs' scores clustering near zero, which makes it challenging to provide high-quality similarity adjustment. Elaboration with example is presented in appendix \ref{sec:main_method_example}.

\subsection{Training Process of OASIS}
\label{sec:method3}
During the training phase of OASIS, we adopt two distinct optimization objectives. The first objective employs the traditional InfoNCE loss function, where the similarity of positive sample pairs within a batch serves as the numerator, and the similarity across all other negative samples within the batch forms the denominator, as follows:
\begin{equation}
    \small
        \mathcal{L}_{ibn} = -\sum_{b}\sum_{i=1}^m\log \Big[\frac{\exp(\text{cos}(h_i, h_i^+)) / \tau}{\sum_{j=1}^N \exp(\text{cos}(h_i, h_j)) / \tau}\Big]
\end{equation}
where \(\tau\) is a temperature hyperparameter, \(b\) stands for the \(b\)-th batch, \(h_i\) and \(h_i^+\) represents the embeddings of a positive pair, and \(h_j\) is the embedding of every sample from the same batch, \(m\) represents the number of positive pairs in \(b\)-th batch, \(N\) is the batch size, and \(cos(·)\) is the cosine similarity.

The second loss function utilizes CoSENT \cite{huang2024cosent}, which uses the order of sample pair similarities within a batch as the objective. It aims to align the predicted rank of sample pair similarities with that of the ground-truth labels. This optimization objective does not focus on the specific similarity values of sample pairs but rather on their relative relationships. For instance, if the ground-truth label indicates that the similarity between pairs \((i,j)\) is greater than that between \((m,n)\), and model predicted similarity is \(s_{mn} > s_{ij}\), this will contribute to the loss; conversely, if the prediction is correct, it will be disregarded. 
The order objective function is as below:
\begin{equation}
    \small
    \begin{split}
        \mathcal{L}_{\text{cos}} &= \log \left[ 1 + \sum_{s_{ij} > s_{mn}} \exp\left({\frac{\cos_{nm} - \cos_{ij}}{\tau}}\right) \right] \\
        & \text{where}\ \cos_{ij} = \cos(h_i, h_j)
    \end{split}
\end{equation}
where \(\tau\) is a temperature hyper-parameter, \(s_{ij}\) is the similarity between embeddings \(h_i\) and \(h_j\), \(s_{mn}\) is the similarity between embeddings \(h_m\) and \(h_n\). \(s_{mn} > s_{ij}\) is relationship from the ranking of training data labels generated from previous step.

Through this approach, the model can learn more nuanced differences between sample pairs and uncover implicit information that may be overlooked by InfoNCE. Essentially, InfoNCE focuses on forming a general embedding for the positive pair, whereas CoSENT concentrates on refining the embedding through relative relationships. 
\begin{equation}
    L = w_1\cdot L_{ibn} + w_2\cdot L_{cos}
\end{equation}
Ultimately, two loss functions above are combined to form the overall optimization objective, with \(w_1\) and \(w_2\) serving as hyper-parameters to balance these objectives. The exact values of all hyper-parameters are specified in appendix \ref{sec:appendix_setting}.

\begin{table*}[h]
\begin{center}
\begin{adjustbox}{width=\textwidth}
\begin{tabular}{lccccccccc}
\toprule
\multirow{2}{*}{\bf Model} & \multirow{2}{*}{\bf CoSQA}  &\multirow{2}{*}{\bf AdvTest} 
&\multicolumn{6}{c}{\bf CSN}\\
\cmidrule(lr){4-10}
& & & Python & Java & JS & PHP & Go & Ruby & Avg \\
\midrule    
\midrule
\multicolumn{10}{c}{\textit{Closed-source Models}}\\
\midrule
OpenAI-ada-002 &$44.23$ &$38.08$ &$68.02$ &$71.49$ &$67.50$ &$60.62$ &$85.63$ &$74.72$ &$71.33$ \\
Text-Embedding-3-Large &$55.38$ &$46.84$ &$70.84$ &$72.92$ &$68.13$ &$59.59$ &$87.64$ &$75.25$ &$72.40$ \\
\midrule
\midrule
\multicolumn{10}{c}{\textit{Open-source Models}}\\
\midrule
CodeBERT &$0.24$ &$0.06$ &$0.05$ &$0.03$ &$0.04$ &$0.02$ &$0.14$ &$0.34$ &$0.10$ \\
GraphCodeBERT &$16.20$ &$5.58$ &$10.37$ &$8.59$ &$7.29$ &$8.07$ &$12.47$ &$20.79$ &$11.26$ \\
UnixCoder &$42.11$ &$27.32$ &$42.17$ &$43.92$ &$40.46$ &$35.21$ &$61.39$ &$55.22$ &$46.39$ \\
CodeSage-large &$47.53$ &$52.67$ &$70.77$ &$70.21$ &$69.50$ &$61.33$ &$83.71$ &$71.92$ &$71.24$ \\
\midrule
OASIS &$\bf55.77$ &$\bf57.27$ &$\bf73.69$ &$\bf73.97$ &$\bf69.80$ &$\bf63.84$ &$\bf88.21$ & $\bf75.47$ &$\bf74.16$  \\

\bottomrule
\end{tabular}
\end{adjustbox}
\end{center}
\caption{Evaluation results (MRR scores) of NL2Code Search in zero-shot setting. 
%The results highlighted in bold represent the best performance. 
Results of open-source baselines are obtained from \citet{zhang2024code} and closed-source by our re-implementation.
OASIS achieves the best results in all columns (in boldface) with significant performance gains compared to all others on average ($p<5\%$).
%Advantage is significant on average with p-values $< 5\%$.
}
\label{table:nl2code}
\end{table*}

\begin{table}[h]
    \centering
        \begin{tabularx}{\columnwidth}{XCr}
            \toprule
            Language &Number &Proportion \% \\
            \midrule    
            \midrule
            Python &$9,138,603$ &$16.98$ \\
            Java &$6,033,337$ &$11.21$ \\
            JavaScript &$8,307,821$ &$15.43$ \\
            PHP &$6,686,344$ &$12.43$ \\
            GO &$8,307,821$ &$15.43$ \\
            TypeScript &$4,690,087$ &$8.71$ \\
            C\# &$5,141,546$ &$9.55$ \\
            Ruby &$4,487,369$ &$8.34$ \\
            C &$1,032,574$ &$1.92$ \\
            \midrule
            Total &$53,825,502$ &$100$ \\
            \bottomrule
        \end{tabularx}
    \caption{Number and Proportion of order-augmented sample pairs in each language of training dataset.}
    \label{table:training data statistics}
\end{table}

\section{Experimental Setup}
In this section, we will provide a detailed elaboration of the experimental setup, including datasets, baselines, and evaluation metrics. An extensive training setting is available in the appendix \ref{sec:evaluation_dataset_appendix}.

\paragraph{Dataset.} Following the experimental setup of CodeSage \cite{zhang2024code}, we utilized the Stack \cite{kocetkov2022stack} dataset as our training data, which is collected from open-source repositories on GitHub. We randomly sampled 140k repositories containing various languages from the Stack dataset. These repositories were subsequently processed using the method described in Section \ref{sec:method}, resulting in 53 million high-quality data across nine languages. The statistics of the dataset are presented in Table \ref{table:training data statistics}. We maintained a comparable quantity across the various languages, demonstrating that our data synthesis method can enhance model performance across diverse languages.

\begin{table*}[h]
\begin{center}
\begin{adjustbox}{width=\textwidth}
\begin{tabular}{lcccccccccc}
\toprule
Model &Python &Java &JS &TS &C\# &C &Ruby &PHP &GO &Avg \\
\midrule    
\midrule
\multicolumn{11}{c}{\textit{Closed-source Models}}\\
\midrule
OpenAI-ada-002  &$35.91$ &$25.13$ &$19.01$ &$21.86$ &$10.17$ &$29.15$ &$40.85$ &$40.47$ &$23.43$ &$27.33$ \\
Text-Embedding-3-Large &$41.51$	&$25.75$	&$22.40$	&$22.45$	&$11.56$	&$32.82$	&$41.70$	&$43.47$	&$21.57$	&$29.25$ \\
\midrule
\midrule
\multicolumn{11}{c}{\textit{Open-source Models}}\\
\midrule
CodeBERT  &$14.40$ &$7.62$ &$5.47$ &$6.05$ &$3.66$ &$5.53$ &$13.55$ &$10.28$ &$6.27$ &$8.09$ \\
GraphCodeBERT  &$19.23$ &$10.78$ &$7.38$ &$8.65$ &$5.54$ &$8.48$ &$19.69$ &$15.67$ &$9.65$ &$11.68$ \\
UnixCoder &$30.77$ &$16.45$ &$21.32$ &$21.95$ &$6.19$ &$15.62$ &$32.33$ &$31.93$ &$13.94$ &$21.17$ \\
CodeSage-large &$46.70$ &$33.13$ &$37.16$ &$41.18$ &$16.81$ &$32.89$ &$54.12$ &$52.13$ &$32.48$ &$38.51$ \\
\midrule
OASIS &$\bf66.27$ &$\bf37.26$ &$\bf47.71$ &$\bf51.15$ &$\bf22.18$ &$\bf49.38$ &$\bf58.60$ &$\bf64.06$ &$\bf34.18$ &$\bf47.87$ \\

\bottomrule
\end{tabular}
\end{adjustbox}
\end{center}
\caption{
Evaluation results (MAP scores) of zero-shot Code2Code Search. 
%The results highlighted in bold represent the best performance. 
Results of open-source baselines are obtained from \citet{zhang2024code} and closed-source by our re-implementation.
OASIS achieves the best results in all columns (in boldface) with significant performance gains compared to all others on average ($p<5\%$).
% MAP score of in-language Code2Code Search in zero-shot setting. The results highlighted in bold represent the global best performance. Results of open-source models are obtained from \citet{zhang2024code}. Results of closed-source model are attained by our implementation. Advantage is significant on average with p-values $< 5\%$.
}
\label{table:code2code}
\end{table*}

For evaluation purposes, we conducted assessments on several widely used code search datasets. The benchmarks were categorized into two types: natural language to code (NL2Code) and code to code (Code2Code) searches. The NL2Code category includes the datasets CoSQA \cite{huang2021cosqa}, AdvTest \cite{lu2021codexglue}, and CodeSearchNet \cite{husain2019codesearchnet}, which is extended with extra candidate codes in GraphCodeBert \cite{guo2020graphcodebert}, named CSN (CodeSearchNet). For the Code2Code section, we employed CodeSage's extended language dataset, which includes additional 6 languages along with the original 3 languages.

\paragraph{Evaluation Metrics.} For the NL2Code tasks, all three datasets employ natural language queries to retrieve code from repositories, where there is only one target code. Consequently, Mean Reciprocal Rank (MRR) is commonly used as the evaluation metric, with higher scores awarded for higher rank of the target code in the retrieval results. In this instance, we adhere to the CodeSage setting by employing an MRR@1000 configuration. For the Code2Code retrieval tasks, as each code query has multiple relevant codes, the Mean Average Precision (MAP) is utilized as the metric for evaluation.

\paragraph{Baselines.} In our study, we conducted comparisons with several prominent code embedding models. Open-source models included CodeBERT \cite{feng2020codebert} and GraphCodeBERT \cite{guo2020graphcodebert}, which utilize masked language modeling (MLM) for pretraining, and UnixCoder \cite{guo2022unixcoder}, which uses contrastive learning. Additionally, we evaluated  CodeSage \cite{zhang2024code}, which have been fine-tuned after pretraining with extensive data. Closed-source models in the comparison comprised OpenAI-Embedding-Ada-002 \cite{openai2023textembeddingada} and OpenAI-Text-Embedding-3-Large \cite{openai2023textembedding3}.

The primary objective of the OASIS method was to propose a novel training approach for code embedding models. To validate the effectiveness of this method, training was not conducted using the dataset included in the benchmark. Instead, training utilized data augmented from open-sourced code from github, which excludes data from the test set. Consequently, many code embedding models that were trained on the CodeSearchNet \cite{husain2019codesearchnet} training set were not considered in this analysis. It should be noted that some of zero-shot performance results for some baselines was sourced from CodeSage \cite{zhang2024code}.

% \paragraph{Implementation Details.} We employed the Qwen2.5-1.5B \cite{qwen2.5} as the backbone model for our training. In terms of hyper-parameters, the learning rate was set at 5e-4, and the temperature \(\tau\) in the optimization objective was configured to 0.05, in alignment with established practices. The weights \(w_1\) and \(w_2\) in the overall loss function were optimized using a grid search approach to ascertain the most effective configuration. The random seed for the training process was set to 3407 \cite{picard2021torch}, adhering to the conclusions presented in a recent paper. $\Delta s$ is set to 10\%.

\section{Experimental Results}
Section \ref{result1} will present the main experimental comparative results, while Section \ref{result2} will be dedicated to ablation studies. Subsequently, in Section \ref{result3}, we will further discuss the OASIS framework.
\subsection{Main Comparison Results}
\label{result1}
In the NL2Code search domain, Table \ref{table:nl2code} demonstrates that OASIS (1.5B) consistently outperforms all open-source baseline models and closed-source models on every language. Compared to the previous open-source state-of-the-art (SOTA) model, Codesage-Large, OASIS achieved a relative improvement of 17.34\% (an absolute increase of 8.24\%) on the CoSQA dataset, and a relative improvement of 8.73\% (an absolute increase of 4.60\%) on the AdvTest dataset. Across all languages in the CodeSearchNet, OASIS surpassed Codesage, with an average performance gain of 4.10\% (2.92\% in absolute terms) across six languages. Notably, OASIS's performance exceeded that of two closed-source models by OpenAI across three datasets. It is important to highlight that although OASIS's similarity labels were generated by Text-Embedding-3-Large, OASIS's performance exceeded that of Text-Embedding-3-Large, further evidencing the efficacy of the method.

In the code-to-code search context, Table \ref{table:code2code} illustrates that OASIS consistently surpasses all baselines in the same-language searches across all nine languages. Specifically, OASIS achieved an average relative improvement of 75.16\% (an absolute increase of 20.54\%) and 63.66\% (an absolute increase of 18.62\%) over the closed-source embedding models OpenAI-ada-002 and Text-Embedding-3-Large, respectively. These improvements are notably significant. Compared to the open-source state-of-the-art model, CodeSage, OASIS registered an average improvement of 24.31\% (9.36\% in absolute terms). Remarkably, in Python, C, Javascript, and PHP, OASIS achieved an increase of over 10\% in absolute MAP score, underscoring its robust performance in these languages.

From the results of the two experiments above, the following observations can be made: 1. OASIS significantly outperforms all the baselines on average. 2. The margin on the Code2Code task (which is more challenging) is larger, indicating that subtle differences are crucial for code semantic understanding. 3. Performance improvements are observed across all languages, supporting the language-agnostic effectiveness of OASIS.

\subsection{Ablation Study}
\label{result2}
OASIS demonstrated robust performance in previous benchmarks, prompting us to conduct ablation experiments to investigate the contributions of different modules. The results are depicted in Table \ref{table:ablation}. Initially, we assessed the impact of 2 loss functions on the performance of OASIS. The experiment indicated that removing either of the loss functions resulted in performance degradation, and combined loss yielded optimal results. However, the order-based optimization objective is more critical than the contrastive optimization objective, which contributes more significantly to the training outcomes.

% thereby confirming that augmenting traditional contrastive learning loss functions with order-augmented data can further enhance the model’s capability to capture code embedding features. However, traditional contrastive loss functions themselves also make a significant contribution; their combined use yields optimal results.

\begin{table}
  \centering
      \begin{tabularx}{\columnwidth}{Xl}
        \toprule
        Model & MRR \\
        \midrule
        \midrule
        OASIS & $\bf69.75$ \\
        OASIS (w/o sim refinement) & $69.15$\\
        \midrule
        \multicolumn{2}{c}{\textit{Objective}}\\
        \midrule
        only order objective & $67.33$\\
        only contrastive objective & $65.49$\\
        \midrule
        \midrule
        \multicolumn{2}{c}{\textit{Selecting Strategy}}\\
        \midrule
        only use AST candidate pair & $69.46$\\
        only use threshold candidate pair & $69.26$\\
        \bottomrule
      \end{tabularx}
  \caption{The ablation study of OASIS on NL2Code benchmarks (average MRR@1000 on 3 datasets).}
  \label{table:ablation}
\end{table}

Secondly, adopting different strategies to adjust similarity labels can produce more precise similarities, thereby enhancing the quality of the training data. Two adjustment strategies focus on different aspects: the threshold selection strategy directly extracts a small number of suspicious pairs, while the AST strategy is employed to filter out a large volume of low-similarity pairs. Both candidate pair selection strategies are equally important and contribute to the performance improvement of OASIS.

Thirdly, Table \ref{table:ablation} also proved that our performance improvements are not solely attributed to LLM-generated docstrings. Our model achieves an NL2Code task performance of 65.49 when using only vanilla InfoNCE loss and the LLM-generated dataset (baseline loss with llm-generated docstring), which is even slightly below the average performance of CodeSage-Large (65.96). Notably, CodeSage’s dataset is derived from docstrings extracted directly from the original code using AST. The slightly lower performance could be attributed to the lack of weight term for negatives in the InfoNCE loss, potentially making the model susceptible to false negative pairs. When incorporating order-based loss on the same dataset, the model achieves a 1.8\% improvement in MRR. Furthermore, the two distinct similarity refinement strategies each contributes additional improvements to the model's overall 4.26\% improvement, which proves the effectiveness of our method does not stem from high-quality LLM-generated docstrings.

Finally, as our model is slightly bigger, we conducted experiments on earlier and smaller-scale models to mitigate the influence of the model's inherent capabilities. The results demonstrate that the performance improvement does not stem from the inherent strength of the backbone model itself but rather from the generalizability of the method. Appendix \ref{sec:small_backbone_model} presents a detailed analysis.

% Utilizing both strategies in tandem not only achieves the best performance but also reduces computational and temporal costs.

% From the results of the ablation experiments, the following observations can be made: 1. The order-based optimization objective is more important compared to the contrastive optimization objective, contributing more significantly to the training outcomes. 2. Both candidate pair selection strategies are equally important and contribute to the performance improvement of OASIS.

\subsection{Further Analysis}
\label{result3}
To provide further insights, we deeply explored how OASIS can generate high-quality embeddings. We probed the effectiveness of OASIS from three perspectives: through analysis of hard subsets, visualization of embeddings, and case studies.

\paragraph{Hard Subset.} To provide a more detailed evaluation of OASIS's performance within the test set, we extracted a hard subset from the CodeSearchNet dataset, specifically from the Python subset, comprising samples where all three models under study exhibited poor performance. These poorly performing samples are defined as instances where, given a query, the target code snippet does not rank first among the retrieved candidate snippets. OASIS utilizes order-augmented data, wherein sample pairs with different similarity labels are separately calculated for distance in comparison to positive sample pairs. Order-augmented approach enables the model to better discern which components contribute to the functionality required by the query. Consequently, the number of poorly performing samples in OASIS is relatively low, and in these samples, the rank of target code in retrieved candidates is generally higher. In the python subset, OASIS's Mean Reciprocal Rank (MRR) significantly surpasses that of CodeSage-Large and Text-Embedding-3-Large, with the improvements of 5.46\% and 5.35\% as shown in Table \ref{table:subset}. Other languages' results can be found in the appendix \ref{sec:ablation_study_detailed}. 

% Additional results for other languages are presented in the appendix.

\begin{table}[b]
  \centering
      \begin{tabularx}{\columnwidth}{Xl}
        \toprule
        Model & MRR \\
        \midrule
        CodeSage-Large & $45.67$\\
        Text-embedding-3-large & $45.78$\\
        OASIS & $\bf 51.13$ \\
        \bottomrule
      \end{tabularx}
  \caption{The Performance on hard samples of CodeSearchNet Python subset (MRR@1000)}
  \label{table:subset}
\end{table}

% \noindent 
\begin{figure}[h]
    % \begin{minipage}[t]{0.57\textwidth}
\centering
\includegraphics[width=\columnwidth]{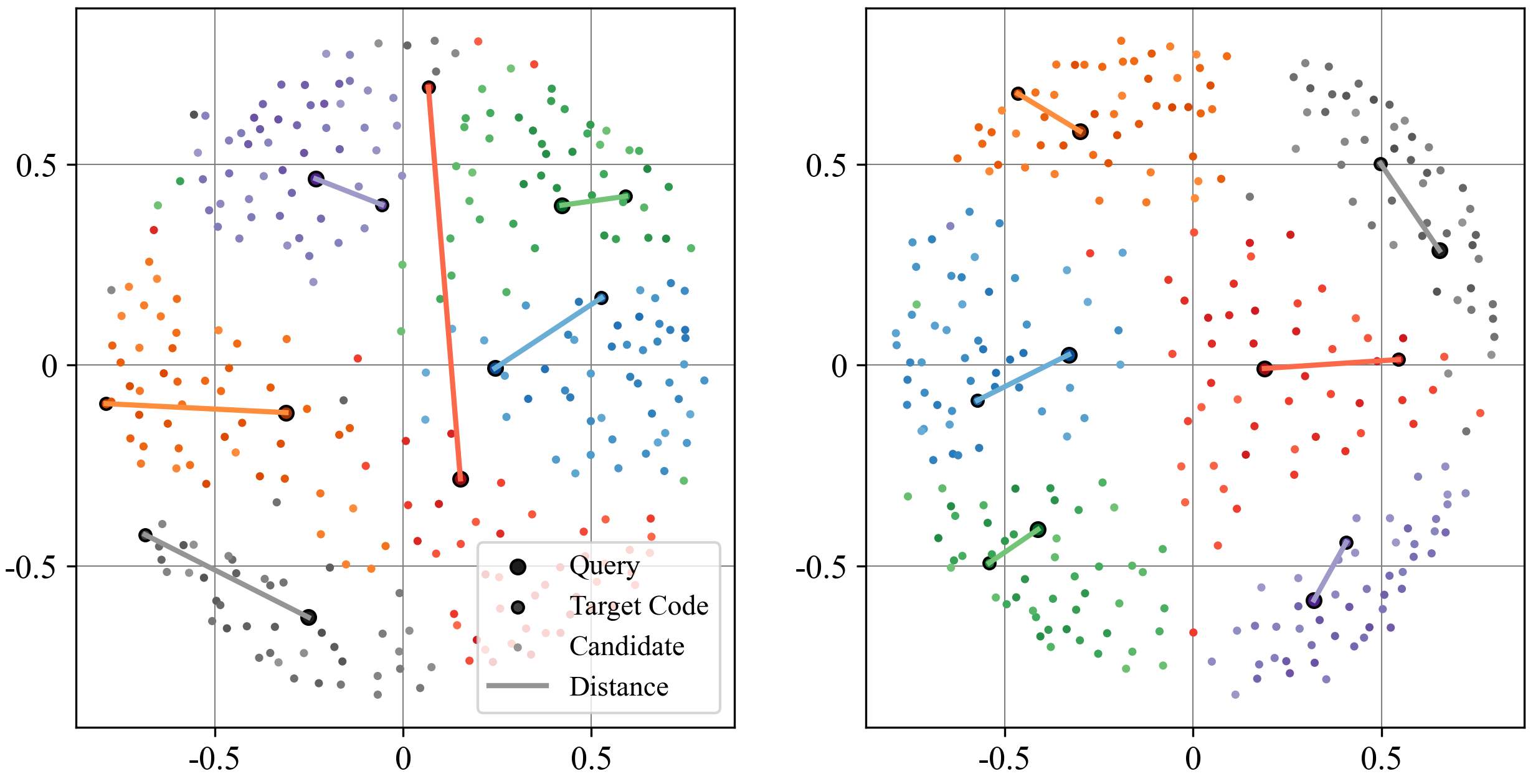}
\caption{Comparison of MDS visualizations between the model without the order-augmented data (left) and OASIS (right), where each colored dot represents a query and its corresponding top 50 candidate codes. Dots with black edges indicate queries and their target codes. Color depth denotes the magnitude of similarity.}
\label{fig:left}
% \end{minipage}
% \hspace{0.03\textwidth}
% \begin{minipage}[t]{0.4\textwidth}
%   \centering
%   \includegraphics[width=\textwidth]{Perf compare distribution.jpg}
%   \caption{Performance Distribution of win tie lose compared to previous models. "Win" indicates the target code appears in a higher rank in results of retrieved candidates.}
%   \label{fig:right}
% \end{minipage}
\end{figure}

\begin{figure}[h]
  \includegraphics[width=\columnwidth]{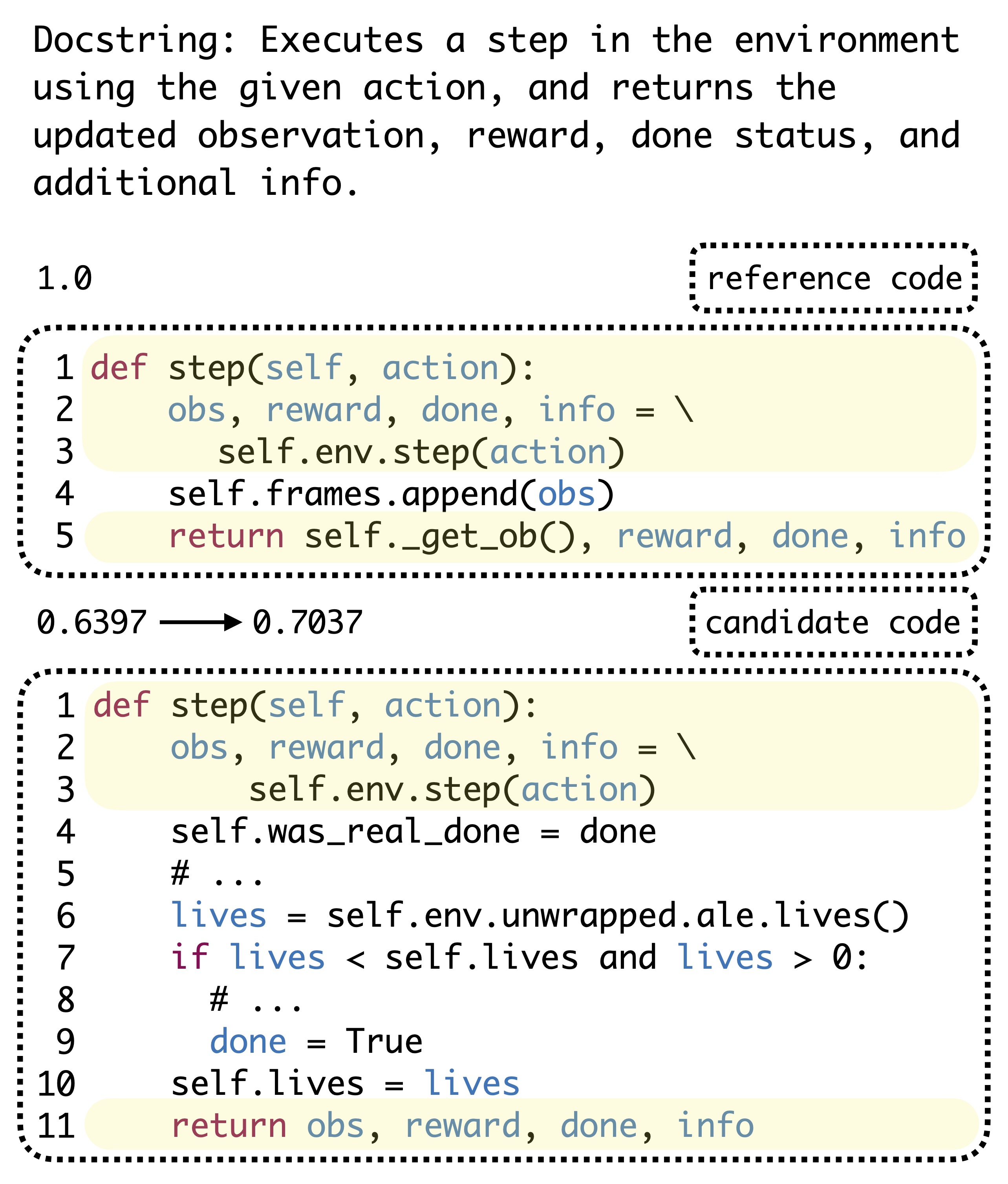}
  \caption{An example of similarity reassignment involves a step function and docstring from a Reinforcement Learning system. Below is a similar candidate function of the same repo. The yellow sections mark functional equivalent parts of the functions. The candidate code also satisfies the description in the docstring, then, the similarity score for the negative pair of the docstring and candidate code was increased by 10\%.}
  \label{fig:case study}
\end{figure}

% Additionally, we compared the performance distribution of OASIS, as depicted in Figure \ref{fig:right}, which illustrates the win-tie-lose distribution relative to previous models. In this context, the terms "win," "tie," and "lose" refer to the ranking of the target code in OASIS's retrieval results compared to other models for a given natural language query. Our findings indicate that OASIS outperformed CodeSage-Large in 24.88\% of the samples, yields same rank in 57.15\% of the samples, and exceeded Text-Embedding3-Large in 23.41\% of the samples, with a tie occurring in 60.39\% of the cases. Overall, compared to CodeSage-Large, the proportion of samples in python subset where OASIS won exceeds those where it lost by 6.91\%, and against Text-Embedding3-Large, the margin is 7.21\%.

\paragraph{Visualization.} Figure \ref{fig:left} presents a MDS visualization, where the model with the same setting except for the absence of the order-augmented data on the left and the OASIS model on the right. 

It is observable that, in comparison with the model without order-augmented data, the embeddings produced by OASIS result in shorter distances between each query and its target code. Each query and all its retrieved candidate codes are distributed within a relatively distinct space, exhibiting less overlap with other queries and candidate codes. This distribution indicates that subtle differences aid in learning in-depth semantics and helps model attend to more essential features, which is particularly important in code's sparser context.

\paragraph{Case Analysis.} To more intuitively understand how the OASIS method assists the model in more effectively learning code embeddings, we also present a case in Figure \ref{fig:case study}. The docstring is generated by LLM for the reference code, forms a positive pair with the reference code, exhibiting a similarity score of $1.0$. This docstring accurately describes the main functionality of the code, which is to perform a single update step in a reinforcement learning system and then return updated reward and other information. The code snippet shown below, drawn from the same repository, forms a negative pair with the docstring, annotated with a similarity score of $0.6397$, which indicates that this candidate code is also highly similar to the docstring. 

The yellow rectangles highlight similar sections between the two code snippets, with the primary difference that the candidate code includes conditional checks on object's lives and some additional processing. The candidate code could also fulfill the description in the docstring, albeit with some variations in the implementation details; thus, the model suggests that candidate code could potentially be closer to the docstring. Based on this assessment, the similarity score of the docstring and candidate code pair was increased from $0.6397$ to $0.7037$ (a $10\%$ increase). This adjustment reduces the distance between this code embedding and the docstring while increasing the distance from other negative pairs with lower similarity in the same batch, thereby achieving more precise code embeddings. Order-augmented data provides an additional, ladder-like signal that enables the model to incrementally approach the code from the query step by step. It instructs the model to focus on the components that fundamentally represent the functionality, as well as on those aspects that subtly distinguish the code from the intent of the query.

\section{Conclusion}
This paper presents OASIS, a novel code embedding model that employs order labels to explore subtle differences among negative NL-code pairs.
It engages a three-step data synthesis method with a hybrid order-based optimization objective, contributing million-scale multi-language training data. 
Extensive evaluation on popular code search benchmarks shows that OASIS pushes SOTA results forward on both NL2Code and Code2Code tasks.

%OASIS is extensively evaluated on popular NL2Code and Code2Code search benchmarks. The experimental results suggest that OASIS outperforms baselines, indicating that OASIS can produce precise and high-quality code embeddings for code search.

\section*{Limitations}
% One limitation arises from constraints in hardware resources, which confined our approach to employing a grid search with only three different delta values during the similarity refinement step to adjust the similarity scores upward for candidate negative pairs. In future work, we plan to explore more refinement strategies and also implement downward adjustments of similarity scores for sample pairs.

Due to constraints in GPU resources, we were unable to extend our method to larger-scale models.
Besides, Our approach is dependent on OpenAI's LLM, and employing alternative open-source LLMs may yield nuanced variations in searching.

\section*{Ethics Considerations}
This study exclusively uses OpenAI's model for research purposes, fully adhering to OpenAI's business terms. We rely on OpenAI's services for data annotation and do not engage in the development or commercialization of competing products. Furthermore, we ensure that no derived models are distributed or shared with third parties, strictly complying with all ethical and legal standards.

\section*{Acknowledgments}
This work is partially supported by the National Natural Science Foundation of China (Grant No. 62372220). Zuchen Gao, Xianming Li, Erxin Yu and Jing Li are substantially supported by Research Grants Council of the Hong Kong Special Administrative Region, China (Project No. PolyU/25200821), the Innovation and Technology Fund (Project No. PRP/047/22FX), and PolyU Internal Fund from RC-DSAI (Project No. 1-CE1E). This work is also partially sponsored by CCF-Kuaishou Large Model Explorer Fund (NO. CCF-KuaiShou 2024013).

\bibliography{custom}

% \clearpage
\appendix

\section{Elaboration of Motivation Case}
\label{sec:motivation_case_elaboration}
The Figure \ref{fig:intro_case} demonstrates the failure to retrieve the correct result, the inability to effectively retrieve relevant results can be attributed to primarily focusing on superficial major features, such as keyword overlaps or approximate semantic matching. For instance, the code snippet ranked first was prioritized due to direct overlaps of keyword “parameter”, while the second-ranked snippet gained precedence by including more keyword “layer” (7, compared to 4 in the third-ranked example). In the top-2 cases, it appears that parameter names were matched to the query to some extent. However, subtle differences that are critical for semantic alignment were overlooked. Specifically, in rank 1, the snippet actually returns the column count or cached output rather than the parameter itself, and rank 2 also returns an output rather than the parameter. By contrast, the target code correctly returns the actual parameter, highlighting the inadequacy of existing methods in capturing these nuanced distinctions.

\section{Analysis of Existing Method for Hard Negative Samples}
\label{sec:hard_negative_analysis}
The methods of leveraging hard negative samples for training can be categorized into two approaches.

The first approach \cite{karpukhin2020dense, gao2021scaling, zhang2021adversarial} involves selecting hard negative pairs by identifying sequences with high similarity to the current sequence based on metrics such as cosine similarity or BM25, while ensuring these sequences are actually dissimilar. These hard negative pairs are then combined with the original positive pairs and randomly selected negatives to construct the training dataset. During training, the loss function does not apply any special treatment to the hard negative samples. The model may be confused and struggle to identify the nuanced differences that decide whether it is the target code.

The second approach \cite{li2022soft, li2023rethinking, zhuang2024not} does not explicitly construct hard negative pairs. Instead, it introduces a weighting mechanism into the InfoNCE loss, where the model assigns lower weights to hard negatives. Essentially, this approach ignores the high similarity of hard negatives and treats them as regular negative samples. Consequently, the observed performance improvement may stem from mitigating the adverse impact of false negatives on the model.

\section{A Detailed Elaboration with Examples for Main Method}
\label{sec:main_method_example}
The case presented in the introduction demonstrates that existing methods struggle to capture subtle differences, particularly for code snippets that appear to match the query but are, in fact, dissimilar. These code snippets along with the query comply with the definition of hard-negative pairs. Traditional approaches for incorporating hard negative pairs suffer from their respective limitations. To address this, in our method, after Step 3.1, we obtain triplet data in the form of 
$(NL, Code, Sim)$, where each natural language (NL) or code query is associated with multiple code snippets. For a given NL query, the dataset contains multiple pairs with different code with varying similarity scores, such as:
\begin{equation}
    \begin{split}
        &(NL_1, code_1, sim1 = 1.0 (0.6)),
        \\&(NL_1, code_2, sim2 = 0.7),
        \\&(NL_1, code_3, sim3 = 0.5),
        \\&(NL_1, code_4, sim4 = 0.2).
    \end{split}
\end{equation}
The similarity of 0.6 is annotated by embedding model, it is manually set to 1.0 during training as it is a positive pair.

While $code_2$ and $code_3$ are not the target code, their similarity scores with remain relatively high compared to unrelated samples. This indicates that $(NL_1, code_2)$ and $(NL_1, code_3)$ constitute hard negative pairs. Unlike traditional methods, our approach has two key features:

\textbf{Preserving Relative Similarity through Order Loss}: Instead of treating hard negative pairs as purely negative samples, the order-based loss leverages the relative similarity among these pairs. It trains the model to preserve the order relationship of similarity scores, capturing the subtle distinctions between them.

\textbf{Progressive Learning with Hard Negatives}: Each NL or code query forms multiple hard negative pairs with varying degrees of similarity, creating a sequence of progressively harder negative samples closer to the positive target. For instance, the model may learn that the dissimilarity between $code_3$ and the target code arises from two minor differences, whereas the dissimilarity between $code_2$ and the target code stems from only one small detail. This progressive learning enables the model to focus on the nuanced variations that distinguish hard negatives from the target code, rather than fully regarding them as irrelevant negatives. By doing so, the proposed framework effectively captures these fine-grained differences and enhances the model's ability to distinguish between similar yet non-identical code snippets.

\subsection{Concrete details of two methods (filtering strategies)}
To fully utilizing hard negative pairs with order-based loss, it is essential to ensure that the similarity labels provided have a high degree of accuracy. However, the labels generated by other embedding models are often suboptimal and require more fine-grained adjustments. To address this, we propose two distinct strategies for selecting pairs that require refinement. Using the earlier example of:
\begin{equation}
    \begin{split}
        &(NL_1, code_1, sim1 = 1.0 (0.6)),
        \\&(NL_1, code_2, sim2 = 0.7),
        \\&(NL_1, code_3, sim3 = 0.5),
        \\&(NL_1, code_4, sim4 = 0.2).
    \end{split}
\end{equation}

The two strategies are as follows:

\textbf{Threshold-Based Suspicious Candidate Filtering}: If, for a given NL-query (e.g., $NL_1$), any negative pair with $NL_1$ exhibits a similarity score that exceeds the original annotated similarity of the positive pair (e.g., $code_2$) or crosses the threshold where the distribution of positive and negative samples overlaps (e.g., $code_3$), such pairs are chosen as suspicious candidate pairs. Because these pairs exhibit a higher similarity compared to the positive pair or compared to the statistical boundary of average positive and negative pairs. This indicates that these negative pairs may require further adjustment.

\textbf{Structural Similarity-Based Filtering}: Even if a negative sample has a low similarity score (e.g., $code_4$ with $sim_4=0.2$), it is also marked as a suspicious candidate pair if its Abstract Syntax Tree (AST) exhibits a certain degree of structural similarity to the AST of the target code (e.g., $code_1$ ). This suggests that structurally similar negative samples may warrant additional adjustment.

The flagged suspicious candidate pairs (e.g., $(NL_1, code_k, sim_k)$) are then input to LLM alongside the positive pair (e.g., $NL_1, code_1$). The model is tasked with determining whether the negative samples (e.g., $code_k$) better address the NL-query (e.g., $NL_1$) in some way. If the model confirms this hypothesis, the original similarity score $sim_k$ is adjusted by applying a small positive offset proportional to a scaling factor $\Delta s$, such that:

\begin{equation}
    sim_k = sim_k * (1+\Delta s)
\end{equation}

This adjustment process refines the similarity scores, leading to improved overall performance. The purpose of both strategies is to identify and flag potentially inaccurate suspicious candidate pairs for further processing and refinement.

\begin{table*}[h]
\begin{center}
\begin{adjustbox}{width=\textwidth}
\begin{tabular}{lccccccccc}
\toprule
\multirow{2}{*}{\bf Model} & \multirow{2}{*}{\bf CoSQA}  &\multirow{2}{*}{\bf AdvTest} 
&\multicolumn{7}{c}{\bf CSN}\\
\cmidrule(lr){4-10}
& & & Python & Java & JS & PHP & Go & Ruby & AVG\\
\midrule
\midrule
Qwen2.5-Coder-1.5B (w/o training) &$0.0186$	&$0.0051$	&$0.0048$	&$0.0039$	&$0.0067$	&$0.0011$	&$0.0034$	&$0.0193$ &$0.0065$  \\
OASIS (codebert-mlm) (125M) &$\bf56.71$	&$\bf42.49$	&$63.32$	&$\bf65.66$	&$\bf60.04$	&$\bf55.82$	&$\bf83.12$	&$\bf67.84$  &$\bf65.97$\\
CodeSage-small (130M) &$49.92$	&$41.28$	&$\bf64.38$	&$63.19$	&$60.01$	&$54.71$	&$77.66$	&$63.20$ &$63.68$ \\
\bottomrule
\end{tabular}
\end{adjustbox}
\end{center}
\caption{Detailed result (MRR score) of NL2Code Search in zero-shot setting. The results highlighted in bold represent the global best performance. OASIS with codebert-mlm backbone achieved the best overall performance.}
\label{table:small_backbone_model}
\end{table*}

\section{Smaller Backbone Model}
\label{sec:small_backbone_model}
The performance margin may source from the capacity and scale of the backbone model, and Qwen2.5-Coder-1.5B is indeed a powerful backbone model and it is slightly bigger than CodeSage-Large (1.3B). To guarantee the generalization of our method, we conducted experiments when training on small-scale and earlier non-decoder backbone model CodeBERT \cite{feng2020codebert}.

Table \ref{table:small_backbone_model} shows the results of NL2Code tasks when using codebert-base-mlm as backbone model. We use the hidden states of cls token as the embedding. It clearly shows that the backbone model Qwen2.5-Coder-1.5B w/o training lacks the capability to perform retrieval tasks. When training with backbone model of codebert, OASIS can still outperform CodeSage-small in the same scale. With a margin of 2\% on MRR on CSN, 6.8\% on CoSQA, and 1.21\% on AdvTest, which proves the effectiveness of our method on different scales of models.

\section{Training Settings}
\label{sec:appendix_setting}

\paragraph{Model and Data.} For training model, the training of OASIS utilized \textit{Qwen/Qwen2.5-Coder-1.5B} \cite{qwen2.5} as the backbone model. For training data, the statistical details of the training dataset are presented in Table \ref{table:training data statistics}, which displays the quantity and proportion of data in different languages within the training dataset after processing through OASIS. Evaluation scripts are available at \url{https://github.com/Zuchen-Gao/OASIS}.

\paragraph{Hyper-Parameter.} In the order-augmented phase, the parameter $K$ was set to $5$. The model utilized for generating similarity scores was the Text-embedding-3-large. Within the threshold strategy for filtering candidate pairs, the threshold $s^*$ was established at $0.4$. Furthermore, in the AST strategy, the filtering threshold was set at $0.25$. During the training of OASIS, the temperature \(\tau\) in the optimization objective was configured to $0.05$, and the weight for the contrastive loss $w_1$ was set at $0.98$, while the weight for the order objective loss $w_2$ was configured at $0.02$. The input length was established at $1024$ tokens, and the pooling strategy employed was last token pooling. A learning rate of $5e-4$ was utilized, with a batch size of $5120$. The random seed for the training process was set to 3407, adhering to the conclusions presented in \citet{picard2021torch}. The results were recorded from the first epoch. In the third step of similarity refinement, it was necessary to adjust the similarity of sample pairs that satisfy the filtering criteria. We utilized grid search to explore performance variations with different values of $\Delta s$ on all of the NL2code validation set. Finally, from three different values tested $(0.05, 0.1, 0.2)$, the optimal $\Delta s$ value of $ 0.1$, which provided the best results in Table \ref{table:delta s result}, was used for the final setting.

\begin{table*}[h]
\begin{center}
\begin{adjustbox}{width=\textwidth}
\begin{tabular}{lccccccccc}
\toprule
\multirow{2}{*}{\bf Model} & \multirow{2}{*}{\bf CoSQA}  &\multirow{2}{*}{\bf AdvTest} 
&\multicolumn{6}{c}{\bf CSN} & \multirow{2}{*}{\bf Avg}\\
\cmidrule(lr){4-9}
& & & Python & Java & JS & PHP & Go & Ruby \\
\midrule
\midrule
OASIS ($\Delta s=0.05$) &$57.14$	&$64.61$	&$73.56$	&$74.77$	&$69.92$	&$64.15$	&$89.66$	&$79.36$	&$71.65$ \\
OASIS ($\Delta s=0.2$) &$\bf57.65$	&$64.73$	&$73.31$	&$74.65$	&$69.62$	&$63.91$	&$89.40$	&$79.33$	&$71.58$ \\
\midrule
OASIS ($\Delta s=0.1$) &$56.96$	&$\bf65.03$	&$\bf73.66$	&$\bf75.01$	&$\bf70.03$	&$\bf64.19$	&$\bf89.68$	&$\bf79.74$	&$\bf71.79$  \\

\bottomrule
\end{tabular}
\end{adjustbox}
\end{center}
\caption{Evaluation result (MRR score) of NL2Code Search in zero-shot setting on \textbf{validation} dataset. The results highlighted in bold represent the global best performance. Best results are acquired when $\Delta s$ is set to 0.1.}
\label{table:delta s result}
\end{table*}

\begin{table*}[h]
\begin{center}

\begin{tabularx}{\textwidth}{XCCCCCCCC}
\toprule
\multirow{2}{*}{Num} & \multicolumn{1}{c}{CoSQA}  &\multicolumn{1}{c}{AdvTest} 
&\multicolumn{6}{c}{CSN}\\
\cmidrule(lr){2-2}
\cmidrule(lr){3-3}
\cmidrule(lr){4-9}
&Python &Python & Python & Java & JS & PHP & Go & Ruby \\
\midrule    
\midrule
Query   &$500$ &$19,210$ &$14,918$ &$10,955$ &$3,291$ &$14,014$ &$8,122$ &$1,261$  \\
Candidate    &$6,268$ &$19,210$ &$43,827$ &$40,347$ &$13,981$ &$52,660$ &$28,120$ &$4,360$  \\
\bottomrule
\end{tabularx}

\begin{tabularx}{\textwidth}{XCCCCCCCCC}
\toprule
Num &Python &Java &JS &TS &C\# &C &Ruby &PHP &GO \\
\midrule    
\midrule
Query  &$15,594$ &$23,530$ &$6,866$ &$3,385$ &$11,952$ &$11,260$ &$11,744$ &$6,782$ &$9,720$  \\
Candidate   &$15,594$ &$23,530$ &$6,866$ &$3,385$ &$11,952$ &$11,260$ &$11,744$ &$6,782$ &$9,720$ \\
\bottomrule
\end{tabularx}

% \vspace{1cm}

\end{center}
\caption{Evaluation benchmark statistics of NL2Code (top) and Code2Code (bottom) Search.}
\label{table:benchmark statistics}
\end{table*}

This result explains why when adjusting similarity judgments, we retain the original positive/negative role to avoid performance degradation caused by fluctuations in LLM outputs. This decision ensures that the labeling quality of the dataset is not entirely dictated by the LLM. As demonstrated in Table \ref{table:delta s result}, when the $\Delta s$ is increased to 0.2—allowing for greater LLM intervention—performance actually deteriorates.

\section{Evaluation Dataset}
\label{sec:evaluation_dataset_appendix}

\begin{table*}[h]
\begin{center}
\begin{adjustbox}{width=\textwidth}
\begin{tabular}{lccccccccc}
\toprule
\multirow{2}{*}{\bf Model} & \multirow{2}{*}{\bf CoSQA}  &\multirow{2}{*}{\bf AdvTest} 
&\multicolumn{6}{c}{\bf CSN} & \multirow{2}{*}{\bf Avg}\\
\cmidrule(lr){4-9}
& & & Python & Java & JS & PHP & Go & Ruby\\
\midrule
\midrule
OASIS (full) &$55.77$ &$57.27$ &$\bf73.69$ &$\bf73.97$ &$\bf69.80$ &$\bf63.84$ &$\bf88.21$ & $\bf75.47$ &$\bf69.75$  \\
OASIS (w/o sim refinement) &$54.87$	&$\bf57.52$	&$73.21$	&$73.69$	&$68.84$	&$62.70$	&$87.77$	&$74.60$	&$69.15$ \\
\midrule
\multicolumn{10}{c}{\textit{Objective}}\\
\midrule
only order objective &$55.24$	&$54.56$	&$71.18$	&$71.32$	&$66.43$	&$59.30$	&$88.06$	&$72.54$	&$67.33$ \\
only contrastive objective &$49.40$	&$54.59$	&$69.55$	&$70.33$	&$65.46$	&$58.86$	&$83.04$	&$72.66$	&$65.49$ \\
\midrule
\midrule
\multicolumn{10}{c}{\textit{Selecting Strategy}}\\
\midrule
only use AST candidate pair &$\bf55.81$	&$57.16$	&$73.29$	&$73.87$	&$69.47$	&$63.43$	&$87.76$	&$74.86$	&$69.46$ \\
only use threshold candidate pair &$55.00$	&$57.14$	&$73.43$	&$73.66$	&$69.02$	&$63.00$	&$87.88$	&$74.92$	&$69.26$ \\

\bottomrule
\end{tabular}
\end{adjustbox}
\end{center}
\caption{Detailed ablation study result (MRR score) of NL2Code Search in zero-shot setting. The results highlighted in bold represent the global best performance. OASIS achieved the best overall average performance.}
\label{table:ablation detail}
\end{table*}

\begin{table*}[h]
\begin{center}
\begin{adjustbox}{width=\textwidth}
\begin{tabular}{lcccccccc}
\toprule
\multirow{2}{*}{\bf Model} & \multirow{2}{*}{\bf CoSQA}  &\multirow{2}{*}{\bf AdvTest} 
&\multicolumn{6}{c}{\bf CSN}\\
\cmidrule(lr){4-9}
& & & Python & Java & JS & PHP & Go & Ruby\\
\midrule
\midrule
OASIS &$\bf49.17$	&$\bf41.77$	&$\bf51.13$	&$\bf49.56$	&$\bf44.45$	&$\bf43.89$	&$\bf63.82$	&$\bf50.74$  \\
CodeSage-Large &$38.72$	&$35.53$	&$45.67$	&$42.11$	&$44.19$	&$39.96$	&$49.87$	&$43.34$ \\
Text-Embedding-3-Large &$48.84$	&$27.61$	&$45.78$	&$47.36$	&$41.64$	&$37.27$	&$62.00$	&$50.18$ \\
\bottomrule
\end{tabular}
\end{adjustbox}
\end{center}
\caption{Detailed result (MRR score) of NL2Code Search in zero-shot setting on hard datasets. The results highlighted in bold represent the global best performance. OASIS achieved the best performance in every language.}
\label{table:Hard Dataset Detail}
\end{table*}

The data statistics for the NL2Code and Code2Code benchmarks are displayed in Table \ref{table:benchmark statistics}.
\paragraph{NL2Code.} The NL2Code task involves using a natural language query to retrieve relevant code snippets. We followed the settings of \cite{zhang2024code} and conducted evaluations across three different benchmarks. CoSQA consists of 500 queries sourced from the web and 6268 candidate entries from CodeSearchNet. CSN is a filtered version of CodeSearchNet, encompassing six languages. AdvTest is an adversarial benchmark processed from python subset of CodeSearchNet, where identifiers have been renamed to obscure semantic information within variables, allowing for the assessment of the model's generalization capabilities.

\paragraph{Code2Code.} The Code2Code task involves using a given code snippet as a query to retrieve all relevant code snippets. We conducted tests using an extended test dataset same to that in \cite{zhang2024code}, which includes the original languages of Python, Java, and Ruby, as well as six additional languages: C, C\#, JavaScript, TypeScript, Go, and PHP. The evaluation setup involves searching within a codebase of the same language, meaning the language of the given query is the same as the language of the codebase being searched.

\section{Detailed Experimental Result}
\label{sec:ablation_study_detailed}
\paragraph{Ablation Study.} Table \ref{table:ablation detail} presents detailed results of the ablation study for Table \ref{table:ablation}, demonstrating that the OASIS model, when incorporating all modules, achieved the best performance overall. The removal of any optimization objective from the model invariably led to a degradation in performance, with the hybrid optimization objective delivering the most superior results. Regarding different label refinement selection strategies, OASIS consistently performs the best across all 6 languages in the CSN, and while there were minor fluctuations in performance on CoSQA and AdvTest, it still maintained equivalent levels of efficacy.

\paragraph{Detailed Hard Dataset Results.} Table \ref{table:Hard Dataset Detail} showcases performance results in other languages, akin to those displayed in Table \ref{table:subset}, within the hard dataset. It is important to note that since the hard dataset does not contain samples where the target code rank is 0, the Mean Reciprocal Rank (MRR) is highly likely to be less than 0.5. The results indicate that across all hard datasets in the NL2code search, OASIS consistently outperforms both CodeSage-Large and Text-Embedding-3-large. This demonstrates that in samples where performance is generally suboptimal across different datasets, the search results of rank of target code by OASIS tends to be superior on the whole.

\begin{table}[t]
    \small
    \centering
        \begin{tabularx}{\columnwidth}{XCCCC}
            \toprule
            VS. &Win &Tie &Lose &Total \\
            \midrule
            % \multicolumn{5}{c}{\textit{CoSQA}}\\  
            % \midrule
            % CS-L &$229$	&$152$	&$119$	&$500$ \\
            % TE3L &$177$	&$170$	&$153$	&$500$ \\
            % \midrule    
            % \midrule
            % \multicolumn{5}{c}{\textit{AdvTest}}\\  
            % \midrule
            % CS-L &$7289$	&$7641$	&$4280$	&$19210$ \\
            % TE3L &$9013$	&$7250$	&$2947$	&$19210$ \\
            % \midrule    
            % \midrule
            \multicolumn{5}{c}{\textit{CSN-Python}}\\
            \midrule    
            CS-L &$3711$	&$8526$	&$2681$	&$14918$ \\
            TE3L &$3492$	&$9009$	&$2417$	&$14918$ \\
            \midrule
            \midrule
            \multicolumn{5}{c}{\textit{CSN-Java}}\\  
            \midrule
            CS-L &$2718$	&$6494$	&$1743$	&$10955$ \\
            TE3L &$2073$	&$7080$	&$1802$	&$10955$ \\
            \midrule    
            \midrule
            \multicolumn{5}{c}{\textit{CSN-JavaScipt}}\\  
            \midrule
            CS-L &$714$	&$1883$	&$694$	&$3291$ \\
            TE3L &$742$	&$1933$	&$616$	&$3291$ \\
            \midrule    
            \midrule
            \multicolumn{5}{c}{\textit{CSN-PHP}}\\  
            \midrule
            CS-L &$4210$	&$6514$	&$3290$	&$14014$ \\
            TE3L &$4431$	&$6802$	&$2781$	&$14014$ \\
            \midrule    
            \midrule
            \multicolumn{5}{c}{\textit{CSN-Go}}\\  
            \midrule
            CS-L &$1375$	&$6059$	&$688$	&$8122$ \\
            TE3L &$848$	&$6538$	&$736$	&$8122$ \\
            \midrule    
            \midrule
            \multicolumn{5}{c}{\textit{CSN-Ruby}}\\  
            \midrule
            CS-L &$315$	&$764$	&$182$	&$1261$ \\
            TE3L &$211$	&$843$	&$207$	&$1261$ \\
            \bottomrule
        \end{tabularx}
    \caption{Detailed comparison of OASIS against other models in 6 languages. CS-L is short for CodeSage-Large, and TE3L is short for Text-Embedding-3-large.}
    \label{table:Detailed Perf compare distribution}
\end{table}

\begin{figure}[t]
  \includegraphics[width=\columnwidth]{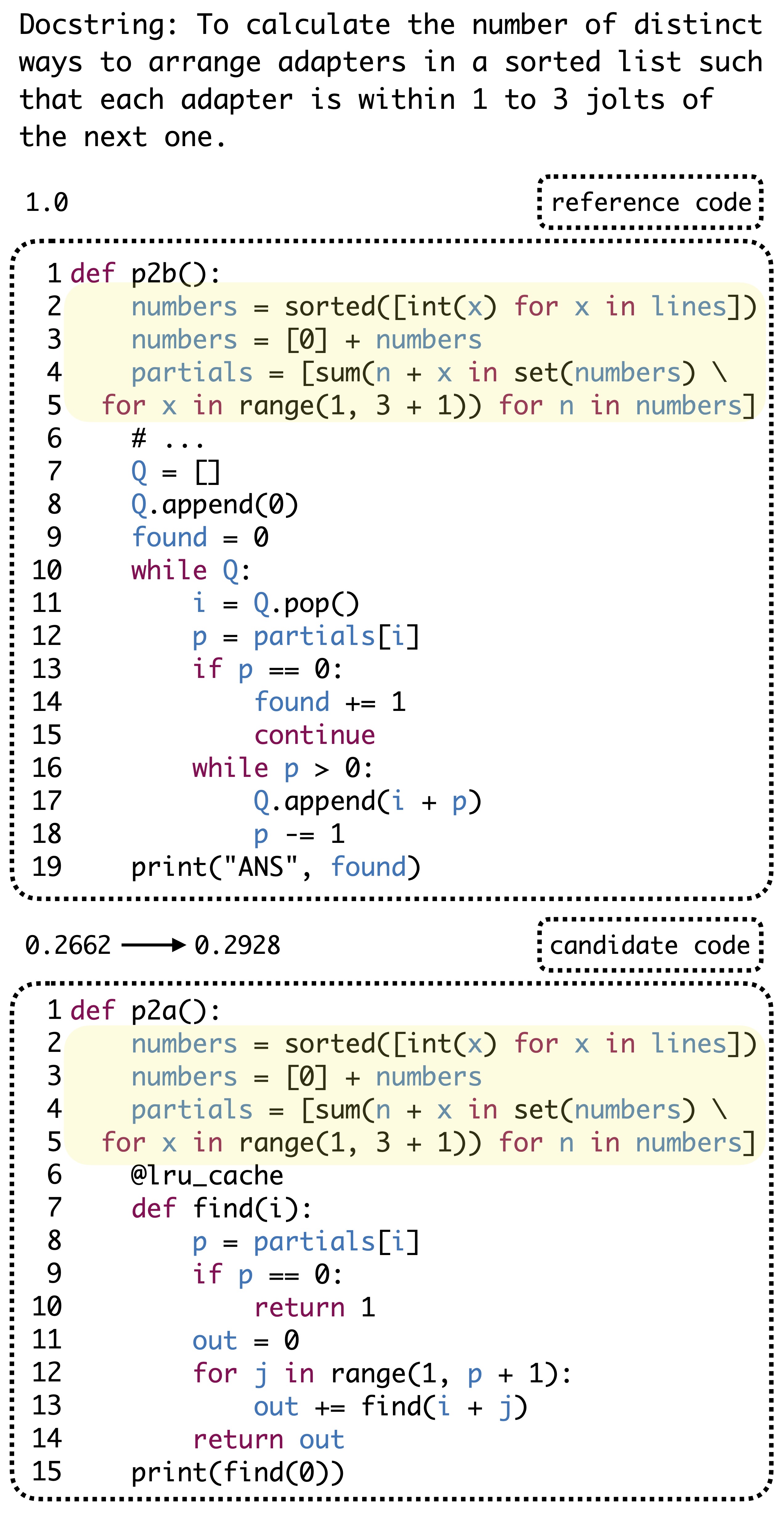}
  \caption{An example of similarity refinement when it triggers AST filtering strategy. The docstring describes a programming challenge from Advent of Code 2020, Day 10 \cite{adventofcode2020}. The use of single-letter variables resulted in a notably low initial similarity score. However, the low edit distance between the ASTs of the two code segments led to the selection of the candidate code. The assessment by LLM confirmed that the candidate code fulfills the requirements specified in the docstring, thereby justifying a refinement of the similarity.}
  \label{fig:case study 3}
\end{figure}

\begin{figure}[t]
  \includegraphics[width=\columnwidth]{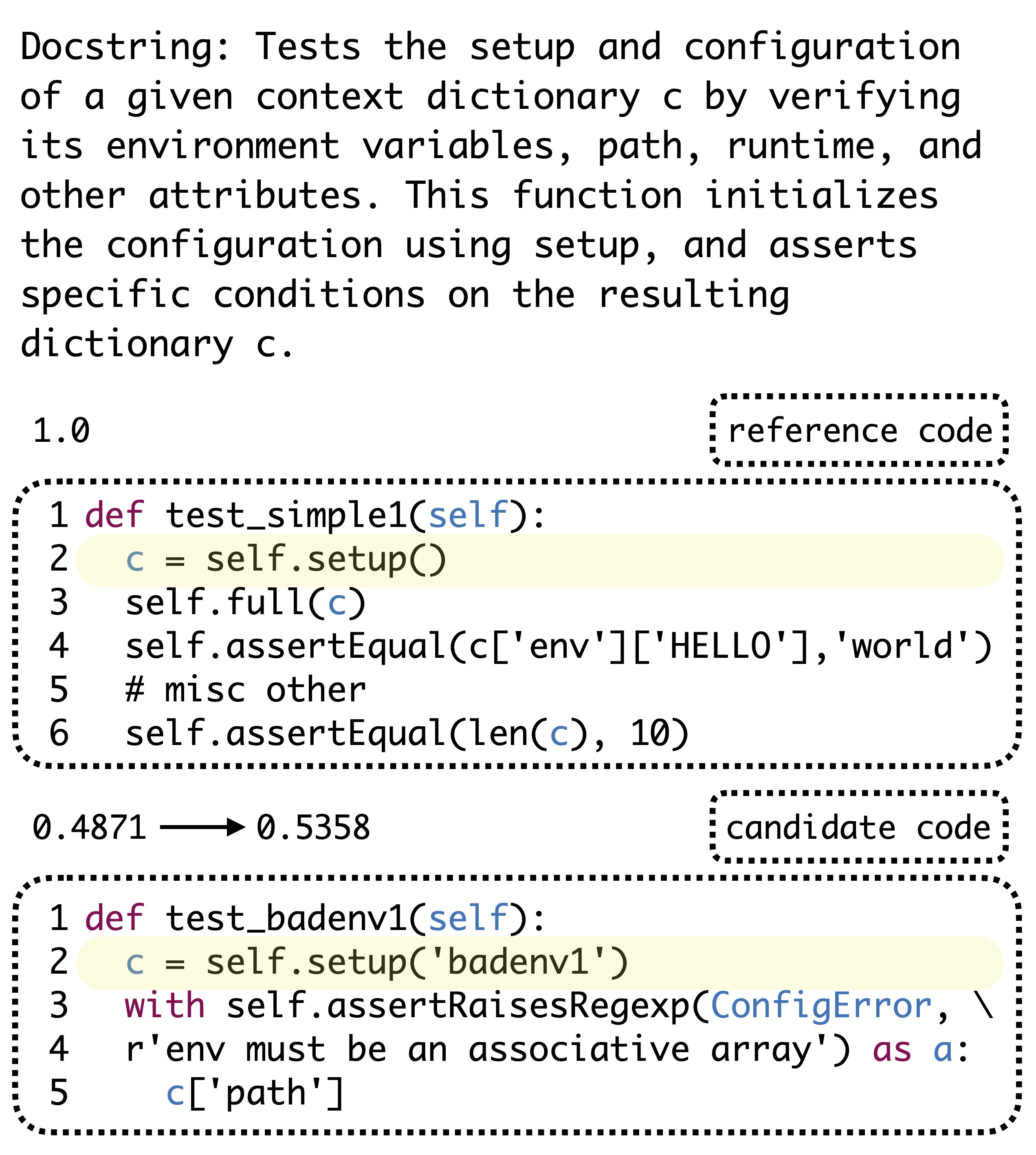}
  \caption{Another example of similarity refinement observed in threshold filtering strategy. The docstring describes the function of the reference code as initializing a configuration followed by performing assertions. The candidate code also accomplishes configuration setup and assertion evaluation. Consequently, adjustments were made to the similarity measurement. }
  \label{fig:case study 2}
\end{figure}

\paragraph{Detailed Comparison.} Table \ref{table:Detailed Perf compare distribution} provides detailed insights into the comparisons for 6 languages of CSN. Consistent with trends observed in Python, OASIS, compared to CodeSage-Large and Text-Embedding-3-large, demonstrates a greater number of wins than losses across all other datasets, indicating a superior overall performance.

\section{Choice of Embedding Model}
The accuracy of code similarity annotations significantly affects the final performance of the model. Different embedding models often exhibit distinct similarity distributions. For example, Text-Embedding-3-Large tends to assign similarity scores around 0.5 for positive pairs and approximately 0.05-0.3 for negative pairs. In contrast, gte-Qwen1.5-7B-instruct assigns similarities around 0.4 for positive pairs and 0.2 for negative pairs, resulting in a narrower similarity range.
However, while the absolute similarity scores vary across different embedding models, the order of similarity scores within a sample group remain the same. For instance, consider a positive pair and three corresponding negative pairs:
\begin{equation}
    \begin{split}
        &(NL_1, code_1, sim1 = 1.0 (0.6)),
        \\&(NL_1, code_2, sim2 = 0.7),
        \\&(NL_1, code_3, sim3 = 0.5),
        \\&(NL_1, code_4, sim4 = 0.2).
    \end{split}
\end{equation}
Here, the similarity score 0.6 for the positive pair is manually adjusted to 1.0 during training. The range of similarity scores (max-min) is 0.5. When switching to another embedding model, the similarity scores for the same group may shrink to a narrower range, such as:
\begin{equation}
    \begin{split}
        &(NL_1, code_1, sim1 = 1.0 (0.5)),
        \\&(NL_1, code_2, sim2 = 0.55),
        \\&(NL_1, code_3, sim3 = 0.4),
        \\&(NL_1, code_4, sim4 = 0.25).
    \end{split}
\end{equation}
In this case, the range is reduced to 0.3. Yet, the relative order of similarity scores within the group remains consistent. Since the loss function depends on the relative ranking of similarity scores rather than their absolute values, using different embedding models may lead to changes in score variance but does not affect the loss computation or the final training outcomes.
To empirically verify that the similarity rankings annotated by different embedding models are approximately consistent, we evaluated two alternative models by re-annotating similarity scores for the same dataset and measuring the rank correlation. Specifically, for each NL, the similarity scores of its corresponding code pairs (e.g., $(NL_1, code_1)$, $(NL_1, code_2)$, etc.) were re-annotated using the alternative models, and the ranking consistency of a group was assessed using the nDCG metric. The evaluation was conducted on a randomly selected subset of 500,000 samples, containing approximately 90,000 unique NL-code groups. The average nDCG scores for the two alternative models are as follows:

$\bullet$ gte-Qwen2-1.5B-instruct: $0.9610$

$\bullet$ e5-mistral-7b-instruct: $0.9914$

These results demonstrate that when using different embedding models for similarity annotation, the relative ranking of similarity scores within the group of pairs with the same NL remain the same. As a result, the training process and final loss remain consistent regardless of the embedding model used for annotation.

\section{More Details}
\paragraph{Computational Cost} For similarity annotation cost, due to separate annotation process, only a rough estimate of the time can be provided. Overall, it took approximately 7 days to complete the similarity annotation for the entire 53M dataset on 4 nodes (100 multi-process each). After the similarity annotations and adjustments were finalized, there was no additional computational overhead during the training phase. This is because order-based loss only relies on the hidden states from the final layer of the model for computation, ensuring that training efficiency remains unaffected.

\paragraph{Details about similarity refinement} For similarity refinement, the threshold-based strategy required approximately 84 hours (3.5 days) to complete, whereas the AST-based strategy took around 60 hours (2.5 days). The refinement process was conducted on a single node with 100 processes. The final training dataset comprises 53,825,502 samples, of which 647,521 samples (1.2\%) were refined using the threshold strategy, 115,689 samples (0.2\%) were refined using the AST strategy, and the 98.6\% of the samples remained unchanged.

\section{More Cases} Figures \ref{fig:case study 3} and \ref{fig:case study 2} illustrate two additional examples of similarity refinement. 
Figure \ref{fig:case study 3} presents a case that triggered the AST filtering strategy. The reference code is derived from a solution to the programming challenge of Advent of Code 2020, Day 10, which aims to identify all possible adapter arrangements. The candidate code, predominantly using single-letter variable names, loses semantic information, resulting in a low initial similarity score of $0.2662$. However, the structural similarity between the candidate and reference codes, due to low AST edit distance, triggered the selection strategy. Upon confirmation through the LMM, the candidate code was an alternative implementation for the challenge, so the similarity was increased from $0.2662$ to $0.2928$.

Figure \ref{fig:case study 2} presents another example where the threshold filtering strategy was employed for similarity refinement. The docstring describes the functionality of the reference code as setting up a configuration and then performing assertions on the fields within this configuration dictionary. The candidate code also completes the configuration setup, as highlighted by the yellow rectangles in the figure, which mark sections of identical functionality. Although the candidate code tests different fields, it also includes assertion, thereby fulfilling the requirements described in the docstring. Consequently, the similarity score was increased from $0.4871$ to $0.5358$.

% This is an appendix.

\end{document}